\documentclass{IOS-Book-Article}

\usepackage{graphicx} 
\usepackage{mathptmx}
\usepackage{comment}
\usepackage{amsmath}
\usepackage{plain}
\usepackage{hyperref}
\usepackage{tikz}
\usepackage{subcaption}
\usepackage{url}

\begin{document}
\begin{frontmatter} 

\title{Geography According to ChatGPT -- How Generative AI Represents and Reasons about Geography}

\author[A]{\fnms{Krzysztof} \snm{Janowicz}},
\author[B]{\fnms{Gengchen} \snm{Mai}},
\author[C]{\fnms{Rui} \snm{Zhu}},
\author[D]{\fnms{Song} \snm{Gao}},
\author[E]{\fnms{Zhangyu} \snm{Wang}},
\author[F]{\fnms{Yingjie} \snm{Hu}}, and 
\author[G]{\fnms{Lauren} \snm{Bennett}}

\address[A]{University of Vienna, Austria}
\address[B]{University of Texas at Austin, USA}
\address[C]{University of Bristol, UK}
\address[D]{University of Wisconsin-Madison, USA}
\address[E]{University of Maine, USA}
\address[F]{University at Buffalo, USA}
\address[G]{Esri, USA}

\begin{abstract}
Understanding how AI will represent and reason about geography should be a key concern for all of us, as the broader public increasingly interacts with spaces and places through these systems. Similarly, in line with the nature of foundation models, our own research often relies on pre-trained models. Hence, \textit{understanding what world AI systems construct} is as important as evaluating their accuracy, including factual recall. To motivate the need for such studies, we provide three illustrative vignettes, i.e., exploratory probes, in the hope that they will spark lively discussions and follow-up work: \textbf{(1) }Do models form strong defaults, and how brittle are model outputs to minute syntactic variations? \textbf{(2)} Can distributional shifts resurface from the composition of individually benign tasks, e.g., when using AI systems to create personas? \textbf{(3)} Do we overlook deeper questions of \textit{understanding} when solely focusing on the ability of systems to recall facts such as geographic principles?
\end{abstract}

\end{frontmatter}

\thispagestyle{empty}
\pagestyle{empty}

\section{Introduction and Motivation}
``\textit{Do we know what AI will know?}'' is among the most profound and consequential questions we can ask today. Finding an answer may take teams from multiple disciplines and methods that we have not yet developed. In this book, we settle with a more concrete but no less relevant question by asking ``\textit{How does geography look according to ChatGPT}''. By geography, we mean how the environment varies across regions, be it cities, mountains, or nations, and how humans interact with this environment. While ChatGPT is among the first and most prominent examples, our investigations will span further and concern (generative) foundation models more broadly. Finally, by asking how geography  \textit{looks according to} such AI systems, we put our focus on \textit{representation}.

Now, intuitively, the notion of representation could pertain to either the study of meaning in an ontological or epistemological sense, or to questions of allocation, i.e., uneven coverage or visibility. The latter case relates to the well-studied notion of \textit{AI biases} \cite{mehrabi2021survey}. For instance, due to representation or coverage bias, certain geographic regions may be overrepresented in model outputs. In contrast, other regions that are less well captured in the training data may be underrepresented. Consequently, the accuracy of model outputs or factual recall may vary geographically \cite{wang2025geobs,manvi2024large,liu2022geoparsing,wu2024torchspatial,moayeri2024worldbench}. However, for the development and deployment of generative AI (GenAI), both notions need to be considered: \textbf{(1)} As will be shown later, GenAI models create strong \textit{defaults}, that is, issues reminiscent of mode collapse where the same prototypical feature is returned over and over again. To give a concrete example, systems such as ChatGPT tend to disproportionately reply with Japan (or, in specific cases, Brazil or Canada) when asked about countries \cite{liu2025operationalizing}. \textbf{(2)} At its core, representation is \textit{never entirely epistemically neutral}, at least not from a cognitivist and constructivist point of view, as there are many ways to construct such representations. The way individuals or societies construct concepts such as \texttt{Poverty} or \texttt{Forest} is a function of their experiences, cultures, age, practices, political affiliations, current context, and so on \cite{couclelis2010ontologies,janowicz2012observation,scheider2023pragmatic}. Although not making active decisions or having experience (yet) themselves, generative AI agents will necessarily select one among those representational alternatives \cite{bender2021dangers}, and, therefore, their outputs will align better with some parts of society and less with others \cite{janowicz2025whose,sorensen2024roadmap}. Even more abstractly, at scale, bias is a key to (human) learning and even survival \cite{gigerenzer2025bias}. However, AI agents are nothing like \textit{artificial humans}; rather, they seem to behave more like collective systems with strongly convergent outputs. Hence, even a minute bias or tendency to favor certain representational alternatives may have significant consequences. Unfortunately, however, these issues and the effects of agentic AI acting as an ecosystem are barely studied and understood, especially not in fields such as geography \cite{van2025opportunities,liu2022review}, urban planning \cite{du2024artificial}, or even the broader geospatial sciences. 

Finally, by focusing on representation, we can go beyond merely studying the ability of AI systems to recall facts or perform (spatial) analysis and instead ask questions such as which regions are described in rich, balanced detail and which are painted with a broad brush. Simply put, we suggest that studying AI \textit{correctness} in isolation is insufficient; what matters equally is how AI represents the world. Whether we like it or not, people and companies \textit{are already using AI on a daily basis} to decide on their vacation destination \cite{xie2024travelplanner,yu2025spatial}, career trajectory, real estate, insurance \cite{rao_et_al21}, or urban planning \cite{du2024artificial,wang2025generative}. In most everyday discourse, \textit{contextual statements matter beyond a more abstract notion of truth} \cite{kuhn2014linked}. Yet, we understand little about the consequences. We already know that AI/ML models have been systematically biased along dimensions such as gender and ethnicity \cite{buolamwini2018gender}; whether they also distort our thinking about and interaction with geography is not well studied yet and a key issue for AI robustness \cite{hendrycks2021many}. Related to robustness is a similar but seemingly less studied notion, namely \textit{brittleness}. 

To give a concrete example, consider this simple yet telling experiment: the aforementioned tendency to form defaults, such as Japan for the category \texttt{Country}, is highly sensitive to variations in prompting. In our experiments with 200 independent queries against GPT-5.1 at temperature 0.3, the system will reply mostly with Japan (168 times) when prompted with `\textit{Name a country, please.}' However, it will start to favor Canada (104 times) for `\textit{Please name a country.}' Despite both prompts being semantically equivalent. It will take a temperature of 1.0 before the system occasionally replies with a third country, namely Brazil. This experiment is not just an illustration of defaults and their strength but also raises a more important question for the interaction with artificial agents: Would we not expect intelligent systems to exhibit distributional stability under mere paraphrasing? 

Finally, before outlining some of these aforementioned representational issues by providing exploratory vignettes, we would like to argue why work in this book relies mostly on studying outputs of \textit{deployed} generative AI systems instead of relying on technologies known from explainable (Geo)AI \cite{hsu2023explainable}, which themselves have methodological limitations  \cite{xing2023challenges}. To make use of an analogy: Cognitive Science studies \textit{what the mind does and how by observing its interaction with the world}, often via the behavior of individual humans or groups. Neuroscience, in contrast, studies \textit{why the brain functions the way it does} by investigating its inner workings and structures such as activation patterns across different regions of the brain during controlled experiments. Both fields complement one another and are at their best when paired. While current AI systems have neither brains nor minds,  by analogy, we can study them from similar perspectives. While explainable AI methods may reveal how and why models (that we can fully access) perform better for certain regions or categories, they cannot reveal the impacts of deployed models on people and places, nor can they be used to understand the interaction between human and artificial agents \textit{in-the-wild}.

\section{Vignette I: Default Strength and Brittleness}
 Before, we introduced the tendency of (generative) foundation models to form \textit{defaults}, that is, strongly favor particular exemplars for a given category. Recently, Liu et al. \cite{liu2026defaults} studied these defaults across about 300 geographic feature types extracted from GeoNames. For instance, among the 11 large language models (LLMs) they considered, 11 would consistently pick the San Diego Zoo when asked to name a zoo, 10 would consistently name the Everglades when prompted for a wetland, and 8 would consistently name mostly Paris as an example of a city. Not only is the geographic footprint of places across all considered feature types biased towards the US and `the West', but the tendency to hold on to these defaults may also be increasing across LLM generations. For example, the older ChatGPT-3.5 seems to produce a more diverse range of outputs than newer models. This is in line with findings from Murthy et al. \cite{murthy2025one}, and Liu et al. \cite{liu2025operationalizing}, and may be due to changes in alignment or a side effect of developers trying to limit hallucinations.
 
 As such, defaults are potentially problematic, e.g., by reducing the geo-diversity of model outputs and thereby causing (mis)representation in downstream tasks such as (over)tourism. Hence, it is worth studying the strength of these defaults. Here we outline an intuitive definition of this phenomenon as an avenue for future studies; see Eq. \ref{eq:ds}.  

 \begin{equation}
   ds(C) \propto min \left\{ T>0 | \exists i \neq c^*: p(i)> \delta \right\} 
   \label{eq:ds}
\end{equation}

\begin{figure}
    \centering
    \includegraphics[width=1\linewidth]{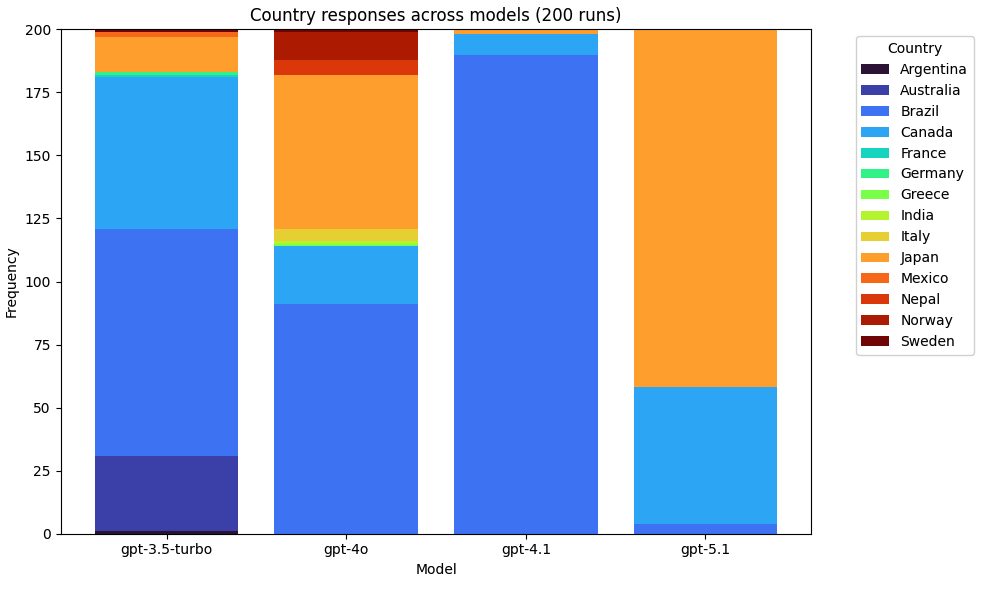}
    \caption{Results for the GPT family for the `Name a country, please.' prompt at a higher temperature of 0.7 (without reasoning for 5.1) to force more varied outputs.}
    \label{fig:ds}
\end{figure}

Simply put, the default strength ($ds$) for a given concept ($C$) is proportional to the minimum model temperature required before we see at least some instances ($i$) of the said concept to be produced as model outputs aside from the default itself ($c^*$), where $\delta$ controls the required empirical frequency of $i$. Returning to our example above, the \texttt{Zoo} default, which predominantly returns the San Diego Zoo, will eventually break as temperature increases, and an increasing fraction of results will be the Schönbrunn Zoo, the Prospect Park Zoo, and so on. The higher this required temperature, the stronger the default. Now, one may argue that such prototypes are simply and exclusively an artifact of training data and that Canada, Everglades, and so forth are merely the most common examples in these datasets. That alone does not readily explain the observed brittleness illustrated above, i.e., why minute changes seem to shift the distribution significantly, nor does it account for differences across models; e.g., GPT-4.1 returns Brazil in 190 of 200 cases, while GPT-5 predominantly favors Japan and outputs Brazil merely 4 times. Figure \ref{fig:ds} shows this convergence \textit{within} a model and simultaneous divergence \textit{across} models. 

As there is no right or wrong answer here, others have recently rightfully asked, `Whose opinions do language models reflect?' \cite{santurkar2023whose}. Clearly, in our collective experience as humans, if we were to ask our friends or colleagues for travel destinations, we would expect a diverse range of responses. Instead, present-day generative AI may present all of us (unknowingly) with the same small set of outcomes, or base what we believe are individualized answers to our prompts on inconsequential syntactic variations in our questions.

\section{Vignette II: Beyond Surface Bias}
A lot of work has gone into debiasing generative AI systems such as text-to-image or large language models \cite{lin2024towards} as well as their underlying data providers such as knowledge graphs \cite{hogan2021knowledge,biswas2023knowledge}. Unfortunately, however, debiasing is not only controversial but may also merely suppresses the deeper underlying issues. First, debiasing itself may lead to misrepresentation as illustrated by Google's Gemini AI producing images of black, Asian, and female German Nazi soldiers in a likely over-correction to increase the diversity of model outputs\footnote{\url{https://www.theguardian.com/technology/2024/feb/28/google-chief-ai-tools-photo-diversity-offended-users}}. Second, debiasing itself is not a neutral task but is based on social norms that may vary by country, region, culture, and so on \cite{janowicz2018debiasing}. Among other challenges, this introduces normative issues similar to those faced by pluralistic alignment\cite{sorensen2024roadmap}, where contextualization becomes a fine line between contextual sensitivity and political influence or even censorship.

In this vignette, we briefly highlight how difficult \textit{deep debiasing} really is and how unexpected distributional shift may re-enter through the \textit{composition} of individually benign tasks. AI agents such as Gemini or ChatGPT will refuse to answer racially loaded questions or those that may imply other forms of discrimination based on gender or religion. These safeguards are effective in general, but they may not necessarily capture potentially deeper or indirect challenges related misalignment or bias. To illustrate this, we explored a two-stage setup to probe such effects, without implying that they would lead to actual discrimination.

\begin{figure}[h]
    \centering
    \includegraphics[width=1\linewidth]{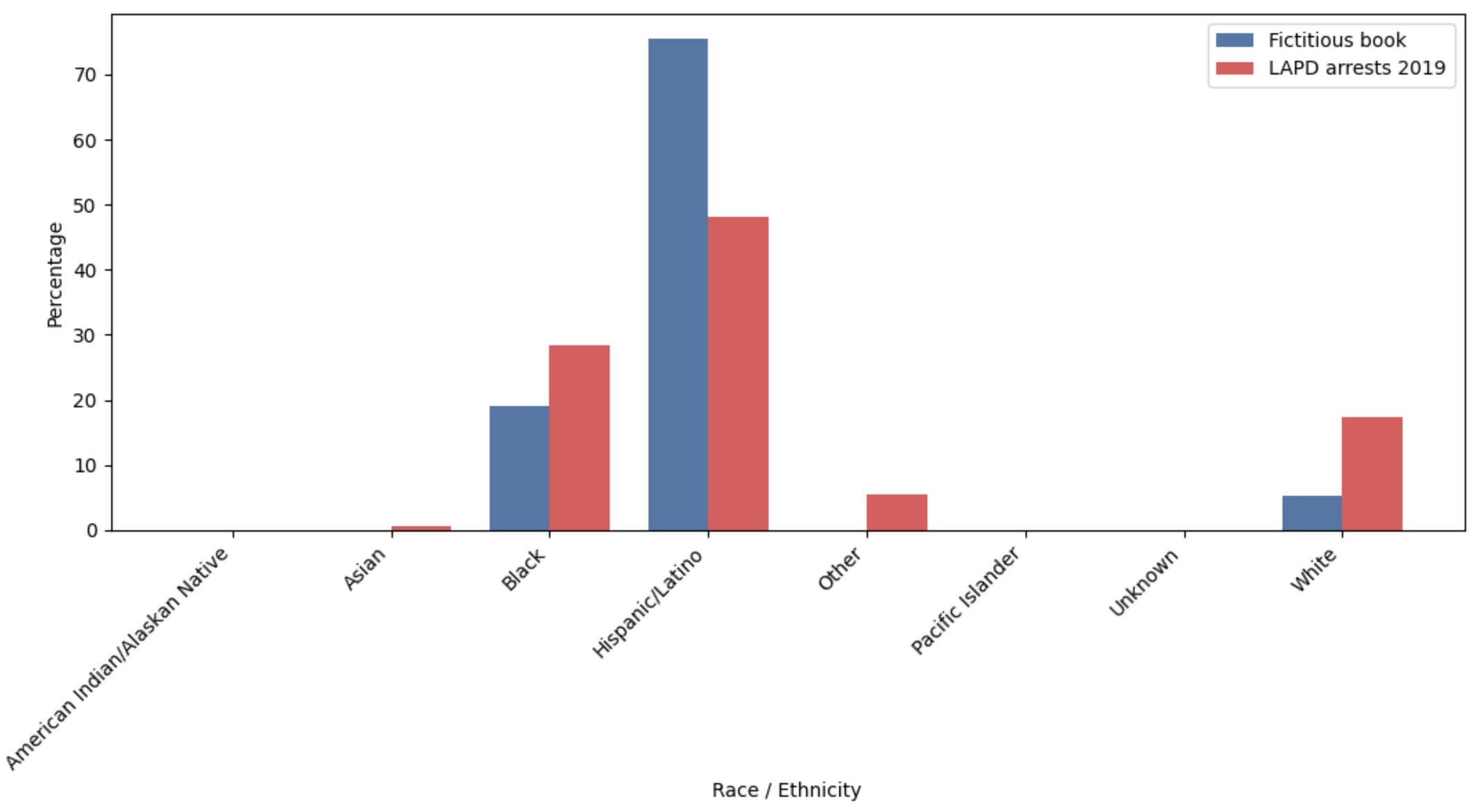}
    \caption{Criminal record labels in the fictitious book population compared to pre-COVID arrest data from the LAPD. Note, the difference between those distributions as such does \textit{not} imply any racial bias.}
    \label{fig:crime}
\end{figure}

\textbf{(1)} We prompted GPT-4o to create 50 realistic personas representative of the greater Los Angeles area. The AI agents compiled a list of people, including names, occupations, ethnicity/race, residency, and age. After running 8 independent queries, we compared the resulting 381 correctly created hypothetical residents with publicly available demographic statistics. The results varied, e.g., the dataset predominantly listed people aged 25-54, and certain occupations were over-represented.\footnote{For ethnicity/race more specifically, the results were: Hispanic or Latino: 35.43\%,  White alone non-Hispanic: 26.77\%,  Black or African American alone: 19.69\%,  Asian alone: 17.59\%, and  Other Race alone: 0.52\%. Hence, this specific generated sample contains more Black or African American and Whites and fewer Hispanics or Latinos than the demographic statistics for 2020.} \textbf{(2)} In a second step, we fed each set of roughly 50 people back to (cold-start) GPT-4o agents, stating that the list of people describes residents of the greater Los Angeles area for a future book. Next, we asked the system to identify which characters had a past criminal record, as we wanted the book to be a crime story. In our runs, the systems assigned criminal roles in patterns that differ from, say, the pre-COVID LA arrest statistics, e.g., by selecting less than 6\% Whites; see Figure \ref{fig:crime}.

As this is a highly sensitive topic that deserves a deeper analysis than a vignette permits, it is important to clarify that we are interested in the composite task of creating personas using AI systems and the observed distributional shift; we explicitly do \textit{not} imply that the observed shift in the sample is a sign of racial bias.  First, crime, offenders, and arrest statistics have their own (racial) biases. Hence, finding a suitable reference distribution is complex. Second, we did not explicitly account for confounding factors such as age, and we did\textit{not} isolate race as the driving causal factor. Third, the study design, such as the prompt phrasing, may also contribute to the results. 

Instead, the key motivation for this vignette is to illustrate how difficult it is to evaluate or control such distributional shifts in generative AI systems, even when explicit safeguards are already in place. As GenAI is already in use across many application areas where artificial personas are created to (implicitly) represent sections of the real population, these issues require follow-up research and guidelines on how best to create such personas. Ideally, symbolic systems such as knowledge graphs may provide reference data for such situations. Closing on a more abstract note, in the future, the cost of validating whether a (Geo)AI model is well-aligned may grow faster than our ability to verify the representational consequences of outputs across geographic contexts and downstream tasks.  

\begin{figure}[h]
    \centering
    \includegraphics[width=1\linewidth]{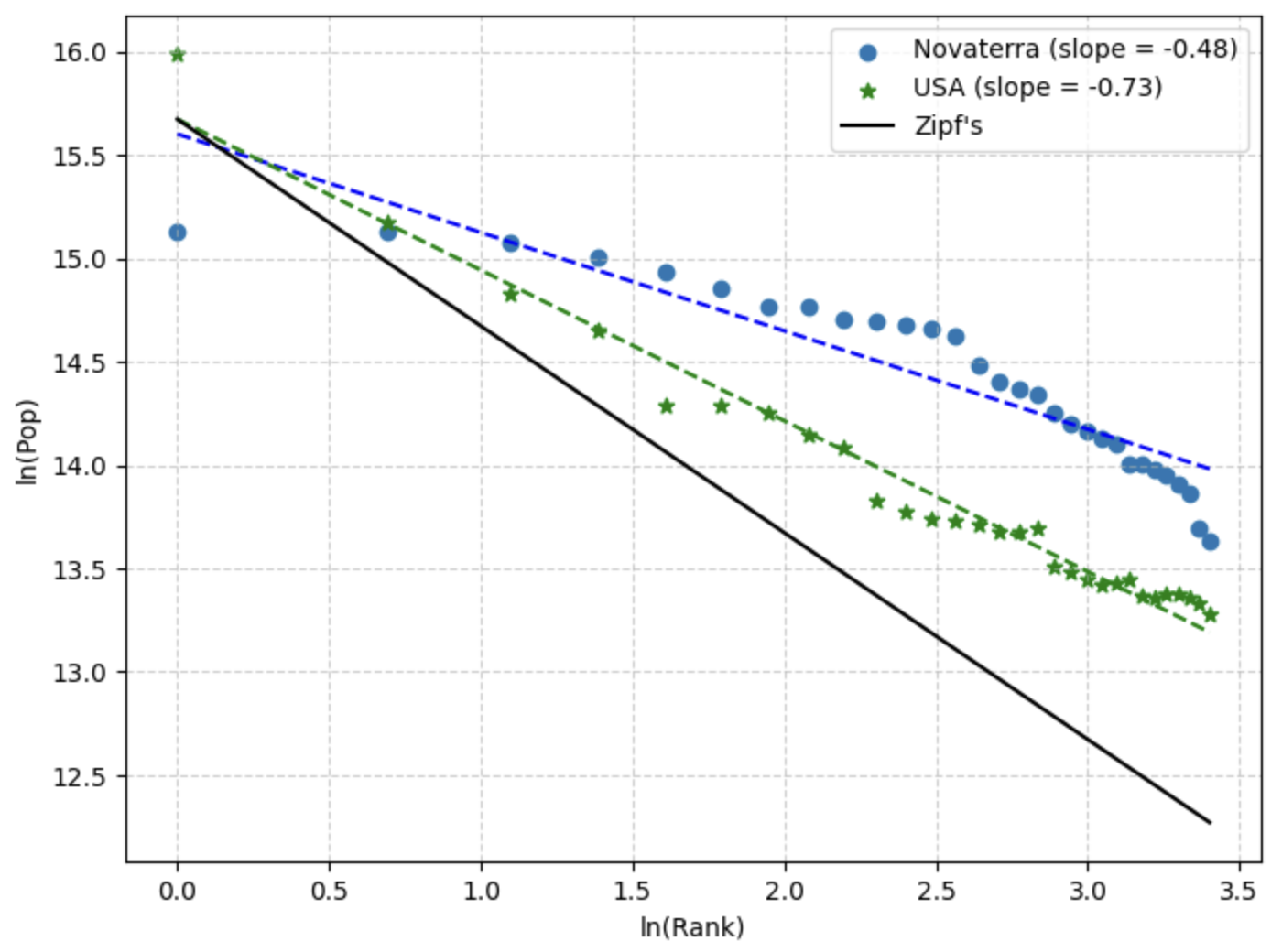}
    \caption{Cities ranked by population: US versus the fictitious island nation of Novaterra.}
    \label{fig:zipf}
\end{figure}

\section{Vignette III: Model Knows, Model Shows}

Another aspect worth considering in the quest to understand how AI systems represent and reason about geographic spaces and places is to surface the difference between what the model can reproduce and what it fully understands. More concretely, there is a notable difference between producing and explaining concepts \textit{on request} and understanding when and how to apply them independently.\footnote{Now, \textit{understanding} is itself a loaded term; what we mean here is an agent's ability to independently apply knowledge to a context without having been explicitly prompted to do so.} Without carefully addressing these differences, we risk confusing the superhuman abilities of AI systems with superhuman intelligence.

To give a concrete example, consider the distribution of city populations. The size distribution of major cities across many nations roughly follows Zipf’s law  (or similar types of distributions). Now, if we prompt AI systems such as Gemini, DeepSeek, or ChatGPT to confirm whether they are aware of this fact, all will provide rich details, examples, and pointers to the literature. However, if we shift towards understanding what such laws imply and how to apply them consistently, the picture changes. 

When asked to imagine a new island nation, say \textit{Novaterra}, at a technical and economic level similar to the US or Japan, a total population of 60 million, and to provide the names and sizes of its 30 largest cities, the models struggled in two notable ways. Across 25 runs (five runs, five models), only two outputs -- both from GPT-5 -- explicitly referenced theories such as the rank-size rule and produced city sizes approximately in line with expectations. Moreover, none of the systems maintained the population constraint, often exceeding 60 million before reaching the rank of 10. Importantly, and in line with our prior argumentation that representation matters beyond accuracy alone, the point is not whether Novaterra's population should have followed Zipf’s law -- many real countries do not -- but whether AI systems can both recall geographic theories and apply them appropriately.  While individual nations vary in their size distributions, taken together, they \textit{do} approximate a rank-size rule. In contrast, however, AI-generated city size distributions did not and varied widely; see Figure \ref{fig:zipf} for a selected example.

To conclude with a broader perspective, this experiment highlights avenues for future work on detecting AI representations, e.g., in geographic misinformation \cite{janowicz2017alternative}. When Li and Goodchild \cite{goodchild2012assuring} became interested in studying the quality of Volunteered Geographic Information (VGI), they turned to an unusual source: the map of the imaginary island of Allestone, drawn by the child prodigy Thomas Williams Malkin. They examined the hand-drawn map, asking whether it would adhere to
well-known geographic principles such as the fractal dimension of the imaginary coastline, 
Horton’s law governing stream bifurcation ratios and Central Place Theory. Astonishingly, the map created by the 5-year-old adhered (approximately) to these principles, suggesting that Malkin had developed a latent understanding of core geographic principles. \textit{Would present-day generative AI systems succeed as well?}

\section{Conclusions and Outlook}
In this chapter, we argued that understanding how generative AI systems represent and reason about the world around us requires going beyond traditional perspectives such as accuracy alone. As work on understanding what AI will (or can) know is still in its infancy, a frequent counterargument is that we do not need to study how GenAI systems represent geography or how these systems \textit{latently} reason about spatial and topological relations, because ultimately it is humans that matter. Unfortunately, however, this view falls short, as generative AI systems are already widely deployed across many downstream tasks, including participatory planning, geospatial intelligence, (tourism) recommender systems, and real estate platforms. Increasingly, the broader public learns about the world through the lens of AI systems. In turn, future AI models may train on geographic representations constructed by earlier models, thereby increasing the risk of model collapse. As we have argued, representing geography is never entirely epistemically neutral. Given that present-day AI systems seem to exhibit a collective (herd-like) behavior, at scale, even minute biases can be consequential. 

We outlined three illustrative vignettes that showcase directions for future GeoAI work: defaults and their brittleness, distributional shifts, and the potential illusion of understanding. Jointly, they point to a broader research agenda on how generative (Geo)AI systems construct, reproduce, and apply geographic knowledge latently and, in turn, how their outputs shape human perception and cognition of the spaces and places around us. While these issues are not entirely novel, there is no widely used set of \textit{practices} how to interact with and study (Geo)AI agents yet. A key motivation for this book was to document early work on how to investigate deployed AI and their interactions \textit{in-the-wild}. Hence, we leave a deeper empirical assessment for future research. We hope that the thematic chapters and surveys provide a fruitful basis for this.

Finally, arguably, the most substantial recent change in the performance of deployed AI systems is not necessarily driven by advancements in their neural architectures alone, but by their ability to decide when to reach out to (more) symbolic methods, such as various forms of retrieval augmented generation (RAG) or the scripting and execution of code. For instance, 2023/24 models would take inputs such as Well-Known Text (WKT) representations of polygons and reason about their topological relations latently \cite{ji2025foundation}. In contrast, 2025 systems load these polygons into a Python script, use libraries such as Shapely for topological reasoning, and return the results. Strictly speaking, by doing so, these systems have entered the realm of neurosymbolic AI, where many open challenges remain unaddressed. At first glance, such neurosymbolic augmentations may reduce visible errors, e.g., by preventing mistakes in latent topological reasoning or spatial hallucination. However, this may merely obscure unresolved latent representation or reasoning issues that may surface unexpectedly in later downstream tasks.

\bibliographystyle{plainurl}
\bibliography{references}
\end{document}